\documentclass{article}

\usepackage{arxiv}

\usepackage[utf8]{inputenc} 
\usepackage[T1]{fontenc}    
\usepackage{hyperref}       
\usepackage{url}            
\usepackage{booktabs}       
\usepackage{amsfonts}       
\usepackage{nicefrac}       
\usepackage{microtype}      
\usepackage{lipsum}		
\usepackage{graphicx}
\usepackage{natbib}
\usepackage{doi}
\usepackage{xcolor}         
\usepackage{bm}
\usepackage{amsmath}
\usepackage{amsthm}
\usepackage{mathtools}
\usepackage{wrapfig}
\usepackage{bm}
\usepackage{multirow}
\usepackage{amsfonts}
\usepackage{thm-restate}
\usepackage{tabularx} 
\usepackage{tablefootnote}

\usepackage{colortbl}
\usepackage{amssymb}

\usepackage{algorithm}
\usepackage{float}
\usepackage{algorithmic}
\usepackage{caption}
\usepackage{xspace}
\newtheorem{theorem}{Theorem}
\newtheorem{lemma}{Lemma}

\usepackage{algorithm} 
\usepackage{algorithmic}
\hypersetup{colorlinks, linkcolor=blue, citecolor=blue, urlcolor=magenta, linktocpage, plainpages=false}

\usepackage{caption}
\newcommand{\ModelName}{Task-Agnostic Gradient Clustered Coreset Selection\xspace}
\newcommand{\ModelNameAbbre}{TAGCOS\xspace}

\title{\ModelNameAbbre: Task-agnostic Gradient Clustered Coreset Selection for Instruction Tuning Data}

\author{Jipeng Zhang$^1$\footnotemark[1]\protect\phantom{\footnotesize 1}, Yaxuan Qin$^{1}$\thanks{\, Equal Contribution. Code are available at the following links: \url{https://github.com/2003pro/TAGCOS}.}\protect\phantom{\footnotesize 1}, \textbf{Renjie Pi}$^1$$^{*}$, Weizhong Zhang$^3$, Rui Pan$^1$,
\textbf{Tong Zhang$^2$}
\\
  $^1$The Hong Kong University of Science and Technology\\
 $^2$University of Illinois Urbana-Champaign \\
 $^3$Fudan University \\
\texttt{\{jzhanggr,rpi,rpan\}@ust.hk,} \texttt{weizhongzhang@fudan.edu.cn,} \\ \texttt{qyxelaine@gmail.com,} 
\texttt{tongzhang@tongzhang-ml.org}
}

\renewcommand{\shorttitle}{}

\hypersetup{
pdftitle={A template for the arxiv style},
pdfsubject={q-bio.NC, q-bio.QM},
pdfauthor={David S.~Hippocampus, Elias D.~Striatum},
pdfkeywords={First keyword, Second keyword, More},
}

\begin{document}
\maketitle
\begin{abstract}
Instruction tuning has achieved unprecedented success in NLP, turning large language models into versatile chatbots. However, the increasing variety and volume of instruction datasets demand significant computational resources. To address this, it is essential to extract a small and highly informative subset (i.e., \textit{Coreset}) that achieves comparable performance to the full dataset. Achieving this goal poses non-trivial challenges: 1) data selection requires accurate data representations that reflect the training samples' quality, 2) considering the diverse nature of instruction datasets, and 3) ensuring the efficiency of the coreset selection algorithm for large models. To address these challenges, we propose \textit{\textbf{T}ask-\textbf{A}gnostic \textbf{G}radient \textbf{C}lustered C\textbf{O}reset \textbf{S}election} (\textbf{\ModelNameAbbre}). Specifically, we leverage sample gradients as the data representations, perform clustering to group similar data, and apply an efficient greedy algorithm for coreset selection. Experimental results show that our algorithm, selecting only 5\% of the data, surpasses other unsupervised methods and achieves performance close to that of the full dataset.
\end{abstract}

\section{Introduction}
Instruction tuning ~\citep{DBLP:conf/iclr/WeiFlan22,DBLP:conf/nips/OuyangInstructGPT22} is the most important strategy for customizing Large Language Models (LLMs) for downstream tasks, which allows them to precisely understand human intentions and accurately generate responses in natural languages. Recently, many existing works \cite{DBLP:conf/nips/WangTULU23} expand the amount and diversity of instructions for instruction tuning to further enhance the LLM's capability.
However, the increased quantity of the dataset also leads to significantly higher computational costs for instruction tuning. Meanwhile, \citet{zhou2023lima} revealed that only 1,000 high-quality, human-created data samples could substantially improve the ability of LLMs to follow instructions, which suggest that there exists severe  redundancy in current instruction datasets, and only a high-quality subset may suffice for achieving promising performance.

To address the above issue, selecting a small, highly informative subset (i.e., coreset) of training samples from the original dataset is a promising solution. This approach ensures that training on the coreset achieves performance comparable to the full dataset while significantly reducing costs. However, coreset selection is challenging as it must not only consider the quality of individual samples, but also their importance within the entire subset. For example, if two high-quality samples are very similar, selecting only one may be sufficient. This global perspective on sample importance is crucial for the quality of the selected subset.

Current methods for coreset selection can be categorized into two main types: 1) Heuristic-based approaches~\citep{DBLP:journals/corr/pplselect,DBLP:journals/corr/ifdscore,DBLP:journals/corr/AlpaGasus,DBLP:journals/corr/instag}, and 2) Optimization-based approaches~\citep{DBLP:conf/nips/BorsosM0bilevelcoreset20, pmlr-v162-zhou22h, gao2023selfguidednoisefreedatageneration, pmlr-v162-zhou22h}. Heuristic-based methods use various heuristic scores to measure sample quality. For example, some assess data sample quality by ranking their corresponding perplexity score~\citep{DBLP:journals/corr/pplselect}, while others score each sample using a powerful LLM~\cite{DBLP:journals/corr/AlpaGasus}. These methods often rely on arbitrary heuristics that may not accurately evaluate sample quality and lack a comprehensive view of sample importance within the entire dataset, resulting in suboptimal performance. Optimization-based methods, on the other hand, typically frame the task as a bi-level optimization problem, requiring repeated optimization of both inner and outer loops. This approach incurs prohibitive costs, especially in the context of large language models (LLMs) that contain billions of parameters. Therefore, a coreset selection method that is applicable for LLMs is yet to be proposed.  

In this paper, to address the above issues, we propose \textit{\textbf{T}ask-\textbf{A}gnostic \textbf{G}radient \textbf{C}lustered C\textbf{O}reset \textbf{S}election} (\textbf{\ModelNameAbbre}), a coreset selection framework designed for LLM that is agnostic of its downstream tasks.
Firstly, we use LLM's gradients as representation for each sample. Compared with representations based on model outputs, gradients effectively captures the information of how each sample affects the optimization direction of the LLM, which is the root cause of the model's final performance. Secondly, to perform  coreset selection under a global view of the entire dataset, we show that coreset selection can be naturally formulated into a Submodular 
Function Maximization (SFM) problem.  Then, noting  that SFM is NP-hard \cite{bach2013learning} and  naive solvers would be impracticable when the dataset size is large, potentially leads to inferior solutions. This urges the development of efficient
approximate optimizer, which is one of the main contributions of this work.  To be precise, we  perform clustering on the gradient features over the dataset to decompose the SFM problem into several small-scaled subproblems to reduce the optimization difficulty.  Lastly, we approximately solve each  SFM subproblems via an efficient greedy approach named optimal matching pursuit (OMP) algorithm to perform  coreset selection independently  in each cluster in a fine-grained manner. This ensures a comprehensive coverage of the selected subset. Our theoretical analysis demonstrates that compared with the methods without our gradient clustering strategy, our method can achieve the comparable accuracy with a significantly smaller sized coreset.

In our experiment, we assessed the effectiveness of our method by selecting data from a combination of 17 popular instruction datasets~\cite{DBLP:conf/nips/WangTULU23,DBLP:journals/corr/tuluv2}, with a total of approximately 1 million data examples. By unsupervisedly selecting 5\% of the original datasets, we obtained great performance on a range of evaluation benchmarks. Additionally, we confirmed the generalization of our method by applying the selected subset to various models.

Our main contributions are as follows:

\begin{itemize}
    \item We verified that gradient features can serve as a good data representation that captures the essential information to measure the quality of instruction data.
    \item We propose \ModelName (\ModelNameAbbre), a coreset selection framework designed for LLM that is agnostic of its downstream tasks.
    \item Our experiment was conducted in a realistic setting, featuring 18 popular instruction datasets that include 1 million varied instruction data points. The practical results convincingly demonstrate the effectiveness of the entire pipeline.
\end{itemize}

\section{Related Work}

\textbf{Instruction Tuning Data.}
Instruction tuning~\citep{DBLP:conf/nips/OuyangInstructGPT22} has achieved unprecedented success in NLP, turning large language models into versatile chatbots~\citep{vicuna2023,alpaca}. Successful instruction tuning requires a powerful pre-trained base model as well as high-quality instruction datasets. For the powerful pre-trained base model, one usually selects a pre-trained LLM with more data and having more parameters, like Mistral~\citep{DBLP:journals/corr/mistral7b}, Llama family models~\citep{DBLP:journals/corr/llama2}. For high-quality instruction datasets part, it is expected that high-quality datasets are diverse and representative enough to adapt the LLM to potential downstream usage. With the development of instruction tuning, there are more and more instruction datasets. Usually, these datasets are either annotated by human or proprietary LLMs. Currently, instruction data generally contains these types: (1) datasets are created by researchers from existing NLP dataset and incorporate an instruction for existing input-output pairs, like Flan~\citep{DBLP:conf/icml/LongpreFlanV223,DBLP:conf/iclr/WeiFlan22}, SuperNI~\citep{DBLP:conf/emnlp/WangSuperNI22}, CoT~\citep{DBLP:conf/nips/WeiCoTData22} and Orca~\citep{DBLP:journals/corr/orca}. (2) open-end text generation, e.g., multi-turn dialogue and instruction following. Several open-end text generation datasets are created by human, like Dolly~\citep{dolly} and Oasst1~\citep{DBLP:conf/nips/KopfOasst23}. Others are generated by proprietary models or human interaction with these models, like Self-instruct~\citep{DBLP:conf/acl/WangSelfInstruct23}, Alpaca~\citep{alpaca}, Sharegpt~\citep{vicuna2023}, Baize~\citep{xu2023baize}, GPT4-Alpaca~\citep{DBLP:journals/corr/gpt4llm} and Unnatural Instructions~\citep{DBLP:conf/acl/HonovichUnaturalInstructionS23}. (3) instructions build for domain-specific skills, like Code-Alpaca~\citep{codealpaca} for code completion. Given such a diverse collection of instruction dataset, the challenge for instruction tuning lies in ensuring the quality of these instructional data samples. \citet{zhou2023lima} revealed that only several high-quality data samples could substantially improve the instruction tuning results. Thus, in this work, we aim to explore an automatic and unsupervised data selection technique to obtain the coreset for these instruction datasets.

\textbf{LLM Data Selection.} Since training LLM still request a lot of resources, data selection is often used for implementing efficient training. Also, several works~\citep{zhou2023lima,DBLP:journals/corr/phi} stress the importance of high-quality data and thus triggered more research works focus on data selection. One popular way to select data samples this is to use an extra LLM to evaluate data samples. \citet{DBLP:journals/corr/AlpaGasus,DBLP:journals/corr/instag} calls ChatGPT API to tag or evaluate the quality of the instruction data. Also, several works~\citep{DBLP:journals/corr/rewarddatasel1,DBLP:journals/corr/rewarddatasel2,DBLP:journals/corr/raft} make use of a reward model to assess the data quality. \citet{DBLP:journals/corr/qurating,liu2024deita} intends to distill the preference of proprietary LLMs to small models for implementing efficient scalable data selection. This line of data selection methods is very expensive and suffers from interpretability. Another line of works focuses on using signals from the model itself to facilitate data evaluation and selection.  ~\citet{DBLP:journals/corr/pplselect, DBLP:journals/corr/superfiltering} make use of perplexity or its variants to determine if a data sample is good or not. \citet{DBLP:journals/corr/less2024,pan2024gradientcluster} use the gradients and influence function to find the data sample that best matches the validation set for downstream tasks evaluation. \citet{DBLP:journals/corr/ifdscore,cao2023instructionmining} develops their own evaluation metric for assessing data samples. Compared to existing data selection works, our work focuses on selecting influential instruction data in a task-agnostic manner, which utilizes LLM gradients as data representation and perform data selection in each cluster of data separately. 

\begin{figure*}[t]
  \includegraphics[width=0.96\linewidth]{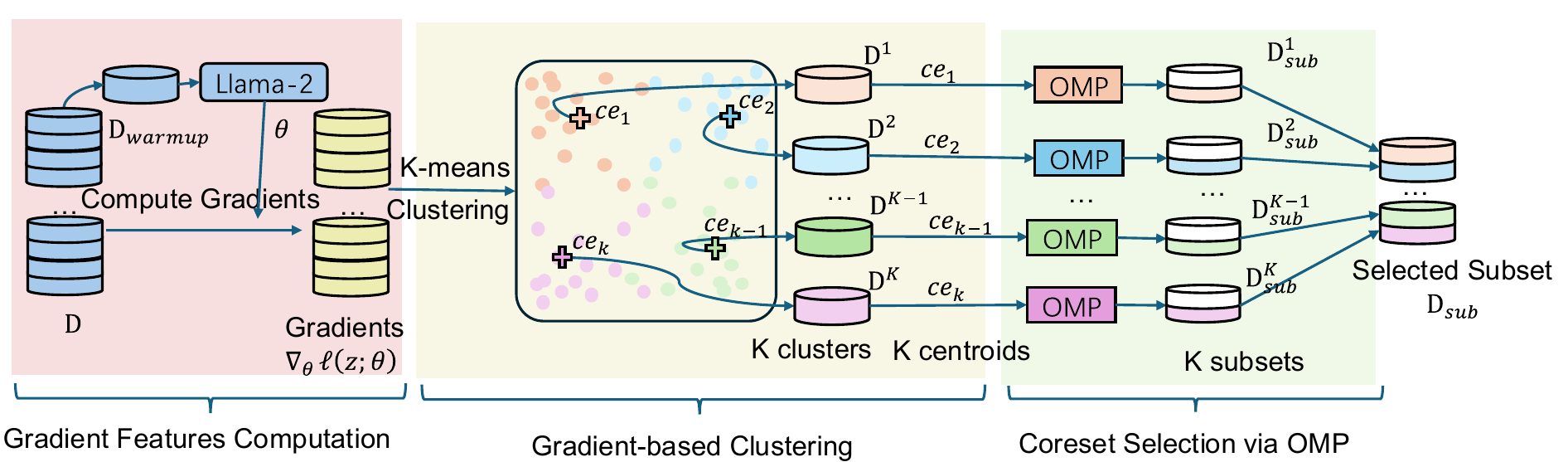} 
  \caption {Illustration of the proposed \textbf{\ModelNameAbbre} pipeline. Our framework consists of three stages: 1) Gradient feature computation, which efficiently derive sample-wise gradients to use as data representation; 2) Gradient-based Clustering, which groups data with high similarity into the same groups; 3) Coreset selection via OMP, which efficiently selects the coresets from each cluster separately in a greedy manner.}
\end{figure*}

\begin{algorithm}[t]
\renewcommand\arraystretch{0.8}
\caption{Coreset Selection }\label{alg:dataselection}
\begin{algorithmic}[1]
\footnotesize
\REQUIRE A pretrained LLM $\theta$, instruction tuning dataset $D = \{ z_i \mid z_i = (s_i, c_i) \}_{i=1}^N$, target subset size $M$, training loss $\ell$, gradient matching error function Err($\cdot$).

\STATE $\theta \leftarrow \text{\textit{FineTune}}(D, \theta)$ \hfill\textit{\# Warm up fine-tune with LoRA}
\STATE $\mathcal{G} \leftarrow \emptyset$
\FOR{each $z_i \in D$}
    \STATE $g_i \leftarrow \nabla_{\theta} \ell(z; \theta)$ \hfill \textit{\# Calculate Sample Gradient}
    \STATE $\mathcal{G} \leftarrow \mathcal{G} \cup \{g_i\}$
\ENDFOR
\STATE \( \{ \mathcal{C}_k \}_{k=1}^K, \{ \mu_k \}_{k=1}^K \leftarrow \text{K-means}(\mathcal{G}, K) \) \hfill \textit{\# Derive clusters and their centroids with K-means}
\STATE $\text{\textit{CoreSet}} \leftarrow \emptyset$
\FOR {each cluster \( \mathcal{C}_k \) with centroid \( \mu_k \)}
    \STATE  $r_k \leftarrow \frac{|\mathcal{C}_k|}{|\mathcal{D}|} \times M$ \hfill\textit{\# Derive subset size in $k^{th}$ cluster}
    \STATE  $\mathcal{C}^{sub}_k, w\leftarrow \text{\textit{OMP}}(\mathcal{C}_k,  r_k, \text{Err($\cdot$)})$ \hfill\textit{\# Derive the subset from $k^{th}$ cluster}
    \STATE $\text{\textit{\textit{CoreSet}}} \leftarrow \text{\textit{\textit{CoreSet}}} \cup \mathcal{C}^{sub}_k$
\ENDFOR

\STATE \textbf{Output:} \textit{CoreSet}
\end{algorithmic}
\end{algorithm}

\section{Method}
To tackle the challenging coreset selection problem for LLM's instruction tuning dataset, we propose \ModelName (\ModelNameAbbre), a task-agnostic coreset selection approach that effectively and efficiently discovers the informative subset from a large instruction tuning dataset. In this section, we first introduce the our formulation of coreset selection, which casts the task into a gradient matching problem. Then, we elaborate the detailed steps for coreset construction.

\textbf{Notation.} Assume we have a pretrained LLM $\theta$ and a giant and diverse instruction dataset $D := \{ (s, c)_{(i)} \}_{i=1}^{N}$, where each data sample $z=(s,c)$ comprises an instruction $s$ and a completion $c$. For each data sample, the loss $\ell(z; \theta)$ is defined as the cross entropy between the prediction distribution $p(\cdot \mid s)$ and the ground truth text response $c$. Since $c$ often contains multiple tokens, $\ell(z; \theta)$ is calculated as the average of the token-wise cross entropy loss across the completion $c$. The notation $\theta_t$ refers to the model checkpoint at step $t$.

\textbf{Problem Formulation.} We first formulate the task into a gradient matching problem, i.e., the average gradient of the selected subset should approximate the gradient of the entire dataset. Intuitively, if the gradient is similar throughout all the training steps, the resulting model parameter should be closed to the model trained with the entire dataset. 

Formally, given a giant and diverse dataset $D$, our goal is to select a subset $D_{sub} \subseteq D$ ($|D_{sub}| < |D|$) containing the most informative training samples. We expect that the gradients produced by the full training dataset $\sum_{z\in D} \nabla_{\theta} \ell(z; \theta)$ can be replaced by the gradients produced by a subset $\sum_{z \in D_{sub}} \nabla_{\theta} \ell(z; \theta)$ with the minimal difference:
\begin{align}
\quad
&\min_{\mathbf{w}, D_{\text{sub}}} \text{Err} \Big( \nabla_{\theta}\mathcal{L}(D;\theta), \frac{1}{\|\mathbf{w}\|_1} \sum_{z \in D_{\text{sub}}} w_z \nabla_{\theta} \ell(z; \theta) \Big) \nonumber \\
&\quad \quad \quad \text{s.t.} \quad D_{\text{sub}} \subseteq D, \quad w_z \geq 0, \quad |D_{sub}|\leq M\label{eq1} 
\end{align}

where $\mathcal{L}(D;\theta) = \frac{1}{|D|} \sum_{z \in D} \nabla_{\theta} \ell(z; \theta)$, $w$ is the subset weight vector, $\|\mathbf{w}\|_1$ is the sum of the absolute values and $\text{Err}(\cdot, \cdot)$ measures the distance between two gradients. 
Note that $w$ could be either continuous, which leads to weighted training on the selected subset, or with discrete values, which reduces to regular training on the coreset. 

However, due to the high diversity of the large-scale instruction tuning dataset, simply conducting selection over the entire dataset potentially causes over-sampling in certain domains and under-sampling in others. To address this, we introduce clustering to ensure balanced sampling. By splitting the dataset into clusters and selecting samples from each cluster, we ensure a more even distribution across different domains.

Overall, as illustrated in algorithm~\ref{alg:dataselection}, the process for coreset construction could be summarized as follows: (1) \textbf{compute the gradient features} $\mathcal{G} = \{ g_i \mid g_i = \nabla_{\theta} \ell(z; \theta) \}_{i=1}^N$. Inspired by \citet{DBLP:journals/corr/less2024}, we compute the low-dimensional approximations of gradient features for each data samples $z$ over the whole dataset $D$; (2) \textbf{perform gradient-based clustering}, we perform $k$-means clustering~\citep{hartigan1979kmeansalgorithm} given the gradients features and get $k$ clusters and corresponding centroids $ce$ for each cluster, which effectively gathers the samples with similar characteristics into one cluster; (3) \textbf{coreset selection via Optimal Matching Pursuit}, we compute the data samples matches best with the centroids in each cluster with an orthogonal matching pursuit algorithm~\citep{DBLP:conf/icml/Killamsettygradmatchomp21}. 

\subsection{Gradient Features Computation}

We perform an efficient gradient feature approximation computation over the entire dataset. To speed up gradient computation, we follow ~\citet{DBLP:journals/corr/less2024} to use LoRA~\citep{DBLP:conf/iclr/HuLoRA22} and random projections~\citep{DBLP:conf/icml/Parkrandomproj23} to reduce the number of dimensions in gradient features. Meanwhile, we propose using checkpoints sampled before convergence to compute gradient features. This is inspired by the fact that the gradient norm calcuated during the warmup phase is significantly larger than checkpoints at convergence. Therefore, these gradient features encapsulate more essential information that reflects how each sample affect the model's updates. The effectiveness of this strategy is verified by results in table~\ref{tab5: gradient checkpoint}.

\textbf{Adam Gradient Computation Function.} The gradients based on Adam optimizer~\citet{DBLP:journals/corr/KingmaAdamB14} can be computed with these steps:

\begin{align}
& \theta^{t+1} - \theta^t = -\eta_t g_i(z, \theta^t) \\
& g_i(z, \theta^t) \triangleq \frac{m^{t+1}}{\sqrt{v^{t+1}} + \epsilon} \\
& m^{t+1} = \left( \beta_1 m^t + (1 - \beta_1) \nabla \ell(z; \theta^t) \right) / (1 - \beta_1^t) \\
& v^{t+1} = \left( \beta_2 v^t + (1 - \beta_2) (\nabla \ell(z; \theta^t))^2 \right) / (1 - \beta_2^t)
\end{align}

where $\beta_1$ and $\beta_2$ are hyperparameters, and $\epsilon$ is a small constant. $g_i(z, \theta^t)$ represents the first-order expansion for the Adam dynamics, requiring model gradients and optimizer states from the training process. Warmup training on a subset of the dataset provides the necessary checkpoints for these computations. As mentioned above, we will sample checkpoints before convergence to provide a more accurate gradient estimation.

\textbf{Warmup Training with LoRA.}  LoRA~\citep{DBLP:conf/iclr/HuLoRA22} is used to reduce the number of trainable parameters and accelerate the inner products in $g_i(z, \theta^t)$. LoRA freezes the pre-trained weights and adds a low-rank adaptor to the selected fully connected layers. We use LoRA to perform instruction tuning on pre-trained base model (e.g., \textsc{Llama}-2-7B) on a random subset $\mathcal{D}_{\text{warmup}} \subseteq \mathcal{D}$ for $N$ epochs, checkpointing the model after each epoch to store $\{\theta_i\}_{i=1}^N$. The gradient when training with LoRA, denoted $\widehat{\nabla} \ell(\cdot; \theta) \in \mathbb{R}^P$, is much lower dimensional than the model itself; for example, in \textsc{Llama}-2-7B, $\widehat{\nabla} \ell(\cdot; \theta)$ is less than 2\% the size of $\theta$. We use $\widehat{\nabla} \ell(\cdot; \theta)$ to compute the Adam update and denote it as $\widehat{g_i}(\cdot, \theta)$.  

\textbf{Projecting the gradients.} Following~\citet{DBLP:journals/corr/less2024}, we also introduce a random project to the LoRA gradients for further reducing the feature dimension. For a given data sample $z$ and model checkpoint $\theta_i$, we can compute a $d$-dimensional projection of the LoRA gradient $\widehat{\nabla} \ell(z; \theta_i) = \Pi^\top \widehat{\nabla} \ell(z; \theta_i)$, with each entry of $\Pi \in \mathbb{R}^{P \times d}$ drawn from a Rademacher distribution~\citep{johnson1984extensions} (i.e., $\Pi_{ij} \sim \mathcal{U}(\{-1, 1\})$). In total, we compute gradient features for each data sample $z$ with $\widetilde{g_i}(z, \cdot) = \Pi^\top \widehat{g_i}(z, \cdot)$.

\subsection{Gradient-based Clustering}
Due to the diversity of instruction tuning dataset, direct sampling over the entire dataset may not cover all the regions, since the training samples from each domain are not evenly distributed. To further improve the effectiveness and robustness of data selection, we divide the entire dataset into several clusters and then perform gradient matching algorithm on each cluster itself. With the gradient features $g_i$ from the above step, we conduct $K$-means clustering on them to assign each data sample into a cluster $ \{\mathcal{C}_k \}_{k=1}^K$. Also, we can obtain cluster centroids $ \{ \mu_k \}_{k=1}^K $ of these clusters during the clustering process, where each centroid shares the dimension with gradient features. 

\subsection{Coreset Selection via Optimal Matching Pursuit}
In each cluster, we hope to get the subset that minimizes the difference between the selected subset and the whole cluster. Instead of doing heuristic selection like selecting all the instances with shortest distance with cluster centroids, we formalize this as an optimization problem and introduce an orthogonal matching pursuit (OMP) algorithm~\citep{DBLP:conf/icml/Killamsettygradmatchomp21,DBLP:journals/corr/ElenbergOMPOri16} to solve it. Similar with equation~\ref{eq1}, our objective is to minimize the difference between selected $D_{sub}^k$ in $k$-th cluster and the whole cluster $D^k$, 
\begin{align}
\text{Err}( \mathbf{w}^k, D_{sub}^k; D^k) \triangleq\notag  \left\|  \sum_{z \in D_{\text{sub}}^k} w^k_z\nabla_{\theta} \ell( z;\theta) - \frac{1}{|D^k|} \sum_{z \in D^k} \nabla_{\theta} \ell(z;\theta) \right\|  \label{eq6} \tag{6}
\end{align}

Considering the regularization coefficient $\lambda$, we can have $\text{Err}_{\lambda}( \mathbf{w}, D_{sub}^k; D^k)$ as:
\begin{align}
\text{Err}_{\lambda}( \mathbf{w}, D_{sub}^k; D^k) \triangleq \text{Err}( \mathbf{w}, D_{sub}^k, D^k) + \lambda \left\| \text{w}\right\|^2.\nonumber
\end{align}

Here, we approximately regard the centroids of each cluster as the average gradients of the whole cluster,
\begin{align}
\frac{1}{|D^k|} \sum_{z \in D^k} \nabla_{\theta} \ell(z;\theta) = ce_k. \tag{7}
\end{align}

We next study the optimization algorithm for solving equation~\ref{eq6}. Our goal is to minimize $\text{Err}_{\lambda}( \mathbf{w}, D_{sub}^k; D^k)$ subject to the constraint $D_{sub}^k : |D_{sub}^k| < d_k$. We can convert this into  maximization problem over the set $D_{sub}^k$ , i.e.,
\begin{align}
&\quad \quad \max_{D_{sub}^k}  F_{\lambda}(D_{sub}^k; D^k), \tag{P-k} \label{prob:P-k}\\
& s.t., |D_{sub}^k|\leq d_k \mbox{ and } D_{sub}^k \subseteq D^k,\nonumber
\end{align}
Here the objective $F_{\lambda}(D_{sub}^k; D^k)$ is defined as 
\begin{align}
\quad
F_{\lambda}(D_{sub}^k; D^k) \triangleq L^k_{\max} - \min_\mathbf{w} \text{Err}_{\lambda}(\mathbf{w}, D_{sub}^k; D^k), \nonumber
\end{align}
where $ L^k_{\max}$ is a constant to make the objective non-negative. 
Note that we minimize $\text{Err}_{\lambda}(D_{sub}^k)$ subject to the constraint $D_{sub}^k : |D_{sub}^k| < d_k$ until $\text{Err}_{\lambda}(D_{sub}^k) < \epsilon$, where $\epsilon$ is the tolerance level and $tk$ is the target num of samples in the selected subset. Note that minimizing $\text{Err}_{\lambda}(D_{sub}^k)$ is equivalent to maximizing $F_{\lambda}(D_{sub}^k)$. Given this, we use OMP to solve this optimization problem, details of OMP are presented in Algorithm~\ref{algorithm2}.

\begin{algorithm}
\caption{OMP}
\begin{algorithmic}

\REQUIRE full dataset $D$, Target subset size $M$, error function Err($\cdot$)
\STATE $D_{sub} \leftarrow \emptyset$
\STATE $r \leftarrow \nabla_{\mathbf{w}} \text{Err}_{\lambda}(\mathbf{w},D_{sub}; D) \big|_{\mathbf{w}=0}$
\WHILE {$|D_{sub}| \leq M$ \textbf{and} $\text{Err}_{\lambda}(\mathbf{w},D_{sub}; D) \geq \epsilon$}
    \STATE $e = \arg\max_j |r_j|$
    \STATE $D_{sub} \leftarrow D_{sub} \cup \{e\}$
    \STATE $\mathbf{w} \leftarrow \arg\min_{\mathbf{w}} \text{Err}_{\lambda}(\mathbf{w},D_{sub}; D)$
    \STATE $r \leftarrow \nabla_{\mathbf{w}} \text{Err}_{\lambda}(\mathbf{w},D_{sub}; D)$
\ENDWHILE
\RETURN $D_{sub}, \mathbf{w}$
\end{algorithmic}\label{algorithm2}
\end{algorithm}

In each cluster $k$, we select data samples that can minimize $\text{Err}_{\lambda}(D_{sub}^k)$ with the above-described OMP algorithm. After finishing the selection on each cluster, we combine the selected subset $D_{sub}^k$ to be $D_{sub}$ and use it to train the target model.

\section{Theoretical Analysis}
In this section, we  analyse the benefits of our gradient clustering in coreset selection. The general conclusion is that coreset selection problem formulated in Problem (\ref{prob:P}) is essentially a Submodular Function Maximization (SFM) problem, which is NP-hard \cite{bach2013learning}. Solving large-scaled submodular function maximization problems is extremely challenging, potentially leads to inferior solution. Our gradient clustering strategy naturally decomposes the original problem into several small scaled problems, significantly reduces the difficulty in optimization, making finding solutions with high-precision possible. The detailed results are presented in the following theorems. These theorems are adapted from the classical analysis on OMP, which can be found in the studies \cite{elenberg2018restricted,wolsey1982analysis}. We adopt them to understand the superiority of our coreset selection approach.

To unify the problems of coreset selection with and without clustering, we extend the problem (\ref{prob:P-k}) as follows:
\begin{align}
&\quad \quad \max_{D_{sub}}  F_{\lambda}(D_{sub}; D), \tag{P} \label{prob:P}\\
& s.t., |D_{sub}|\leq M \mbox{ and } D_{sub} \subseteq D,\nonumber
\end{align}
where $D_{sub}$ and $D$ are the coreset and the full dataset, respectively. $c$ is the constant to control the coreset size.  

\begin{lemma}
    If the coreset size $|D_{sub}|  \leq c$ and $\max_{z \in D} \| \nabla_{\theta} \ell( z;\theta)\|_2 \leq G$, then $F_{\lambda}(D_{sub}; D)$ is $\gamma_D$-weakly submodular with $\gamma_D = \frac{\lambda}{\lambda + MG^2}$. 
\end{lemma}
\begin{theorem}
    If $\max_{z \in D} \| \nabla_{\theta} \ell( z;\theta)\|_2 \leq G$ and $\max_{z \in D^k} \| \nabla_{\theta} \ell( z;\theta)\|_2 \leq G_k$  for cluster $k$. Let $D_{sub}^*$ and $D_{sub}^{k*}$ be the optima of Problems \ref{prob:P} and $\ref{prob:P-k}$, with $k=1,\ldots, K$.  Then, the followings hold:
    \begin{itemize}
        \item[\textup{(i)}] For problem (\ref{prob:P}), OPM runs with stopping criteria $F_{\lambda}(D_{sub}; D) \leq \epsilon$ achieves set $D_{sub}$ with $|D_{sub}|\leq \frac{|D_{sub}^{*}|}{\gamma_D} \log (\frac{L_{max}}{\epsilon})$.
        \item[\textup{(ii)}] For problem (\ref{prob:P-k}), OPM runs with stopping critia $F_{\lambda}(D_{sub}^k; D^k) \leq \epsilon_k$ achieves set $D_{sub}^k$ with $|D_{sub}^k|\leq \frac{|D_{sub}^{k*}|}{\gamma_{D^k}} \log (\frac{L_{max}^k}{\epsilon_k})$.
    \end{itemize}
\end{theorem}
 Since  $\gamma_D = \frac{\lambda}{\lambda + MG^2}$ and $\gamma_{D^k} = \frac{\lambda}{\lambda + d_kG_k^2}$ with $M = \sum_{k=1}^K d_k$,   it can be expected that  $\gamma_D \ll  \gamma_{D^k} $. Noting that a proper clustering method would make $D_{sub}^* \approx \cup_{k=1}^K D_{sub}^{k*}$ and it is reasnonable to set  $\frac{L_{max}^k}{\epsilon_k} \approx \frac{L_{max}}{\epsilon}$ to ensure comparable precisions.  Thus the above theorem demonstrates that
$$\sum_{k=1}^K\frac{|D_{sub}^{k*}|}{\gamma_{D^k}} \log (\frac{L_{max}^k}{\epsilon_k}) \ll \frac{|D_{sub}^{*}|}{\gamma_D} \log (\frac{L_{max}}{\epsilon}).$$ 
That is, to achieve comparable accuracy, the union of the coreset selected from each cluster can be much smaller than that from the whole datasets, which verifies the benefits of gradient clustering. This is also consistent with our experimental observation. i.e., the running time of OMP without gradient clustering is significantly longer than that with gradient clustering. 
\section{Experiment}
In this section, we conduct experiments to answer the following research questions:
\begin{itemize}
\item Does TAGCOS achieve superior performance over other unsupervised selection methods? (Table~\ref{tab1:main_results})
\item How effective is the generalization of TAGCOS, and can it be transferred to different models? (Table~\ref{tab2:generalization})
\item What is the best configuration for TAGCOS, including the selection proportion, the number of clusters, and the selection of gradient checkpoints?~(Table~\ref{tab3:selection proportion}, Table~\ref{tab4: cluster number}, Table~\ref{tab5: gradient checkpoint})
\end{itemize}

\subsection{Setup}

\textbf{Datasets.} To illustrate that TAGCOS is task agnostic, we chose diverse tasks for both training and evaluation. For the training set, we combined 17 popular instruction datasets totaling 1,068,549 examples, following \citet{DBLP:conf/nips/WangTULU23,DBLP:journals/corr/tuluv2}. These datasets vary in format and reasoning tasks, with annotations by humans or the OpenAI API. For details, please refer to Appendix.

For evaluation, we selected \textbf{TydiQA}~\citep{clark2020tydiqa2}, \textbf{MMLU}~\citep{hendrycks2020measuringMMLU}, and \textbf{BBH}~\cite{suzgun2022challengingbbh}. \textbf{TydiQA} is a multilingual QA benchmark covering 11 languages, requiring models to extract answers from passages given a question. F1 is used to as the evaluation metric here. \textbf{MMLU} features multiple-choice questions across 57 subjects, from elementary to professional levels. It asks LLM to select a single correct answer given several options. Accuracy is used as the metric here. \textbf{BBH} includes 23 challenging tasks from Big-Bench, testing general reasoning skills.

\noindent \textbf{Implementation Details.} Following \citet{DBLP:journals/corr/less2024}, we performed warmup training on a randomly selected 5\% of the dataset for 4 epochs and computed 8192-dimensional gradient features on the full dataset $D$. The learning rate for warmup training was set to 2e-5, with a batch size of 32. Using these gradient features, we selected 5\% of the original dataset using our selection methods, totaling approximately 53,427 samples. We used 100 clusters for $K$-means clustering and set the OMP algorithm tolerance at 0.01. After obtaining the subset, we fine-tuned the Llama-2-7B~\citep{DBLP:journals/corr/llama2} and Mistral-7B~\citep{DBLP:journals/corr/mistral7b} models using LoRA~\citep{DBLP:conf/iclr/HuLoRA22} to reduce memory usage. For LoRA training, we used the AdamW optimizer with a learning rate of 2e-5 and 4 epochs. The context length was set to 1,024, with a batch size of 32.

\begin{table*}
\centering
\begin{tabular}{l|c|c|c|c}
\hline
\textbf{} & \textbf{TydiQA } & \textbf{MMLU } & \textbf{BBH} & \textbf{average} \\ \hline
Uniform  
& 52.08  & 46.9  & 41.39 & 46.79  \\
Hardest 
& 51.58 & 45.68 & 38.15 & 45.13 \\
Perplexity
& 51.66 & 46.89 & 40.74 & 46.43 \\
$\text{K-Center}_{BERT}$
& 50.05  & 47.16 & 39.91 & 45.7 \\ 
$\text{K-Center}_{Llama}$
& 52.72 & 46.07 &  39.07 & 45.95\\
 $\text{K-Center}_{Grad}$
& 38.83  & 48.73  & 41.48  & 43.01 \\ 
OMP 
& 53.64  & 46.10  & 40.47  & 46.82  \\  
\hline
TAGCOS  
& 52.78  & 48.01 & 44.26  & 48.35 \\ \hline
\end{tabular} 
\vspace{0.2cm}
\caption{Experimental results on selecting a mix of 17 instruction datasets. The evaluations are performed on the TydiQA, MMLU, and Big Bench Hard (BBH) datasets. All results are based on 5\% data samples selected by the corresponding methods and trained on Llama-2 7B models.}
\label{tab1:main_results}
\end{table*}

\subsection{Experimental Results}

\textbf{Baseline.} The main experiment results are presented in Table~\ref{tab1:main_results}. Several baselines were considered for comparison:  (1) \textbf{Uniform}: randomly selecting the data samples from the original dataset. (2) \textbf{Hardest Sampling}: select the data samples with the highest perplexity. (3) \textbf{Perplexity Sampling}~\citep{DBLP:journals/corr/pplselect,DBLP:journals/coling/Marcusperpelxity94}: select the data samples with the lowest perplexity. (4) \textbf{K-Center-Greedy with different representations}~\citep{DBLP:journals/corr/kcentergreedy}: converting instruction data into embedding vectors, performing $K$-means clustering, and selecting samples by iteratively choosing the one closest to the cluster center among the remaining instances. Here, we consider 3 different embedding spaces, BERT~\citep{DBLP:conf/emnlp/Reimerssentencebert19}, Llama~\citep{DBLP:journals/corr/llama2} and Gradient. We denote them as $\text{K-Center}_{BERT}$, $\text{K-Center}_{Llama}$ and $\text{K-Center}_{Grad}$. (5) \textbf{OMP}~\citep{DBLP:conf/icml/Killamsettygradmatchomp21}: using the OMP algorithm over the entire dataset, with the mean gradient feature across the dataset as the matching target.

\textbf{Main Experiments.} TAGCOS achieves the best performance across all tasks, confirming its efficacy in data selection for instruction tuning. TAGCOS is the only baseline that consistently performs well. Although $\text{K-Center}_{Grad}$ excels on the MMLU benchmark, it fails on TydiQA and is equivalent to uniform sampling on BBH, underscoring TAGCOS's robustness.

\textbf{Effectiveness of each Component in TAGCOS .} The key difference between TAGCOS and $\text{K-Center}_{Grad}$ lies in their selection mechanisms. While $K$-means clustering on gradient features can achieve strong results on individual benchmarks, it is insufficient for consistent overall performance. This further demonstrates the effectiveness of the OMP coreset selection algorithm. Compared to OMP, which does not use clustering, TAGCOS delivers better results. This reinforces our perspective that clustering is essential for managing the diversity in instruction datasets.

\textbf{Gradient Features vs. Other Embeddings.} We evaluated the K-Center algorithm with various data representation schemes, including BERT, Llama, and Gradient. In the absence of a selection mechanism, Llama embeddings, which utilize the last token's hidden representation from the last layer, showed the best results. We attribute this to the closer alignment of Llama features with decoder-only LLM behavior. Additionally, gradient features require an appropriate selection mechanism to exhibit their full potential.

\subsection{Ablation Study and Analysis}

\textbf{Performance of TAGCOS on Different Models.} Table~\ref{tab2:generalization} demonstrates that the dataset generated by the Llama-2-7B model can be effectively utilized to train a superior Mistral-7B instruction model. By leveraging the datasets selected by TAGCOS on the Llama-2-7B model, the trained Mistral-7B model shows significant improvements over uniform selection methods, consistently outperforming its counterparts. This highlights TAGCOS's ability to identify transferrable and valuable data samples, indicating its potential for future proxy data selection tasks.

\begin{table}[ht]
\centering

\begin{tabular}{l|c|c|c|c}
\hline
\textbf{} & \textbf{TydiQA} & \textbf{MMLU} & \textbf{BBH} & \textbf{Average} \\ \hline
\multicolumn{5}{c}{\textbf{Llama-2 7B}} \\ \hline
Uniform  
& 52.08  & 46.9  & 41.39 & 46.79  \\
TAGCOS
& 52.78  & 48.01 & 44.26  & 48.35 \\ \hline
\multicolumn{5}{c}{\textbf{Mistral 7B}} \\ \hline
Uniform  
& 57.59    & 61.34     & 56.48     & 58.47     \\
TAGCOS 
& 61.49     & 61.79     & 57.87     & 60.38     \\ \hline
\end{tabular}
\vspace{0.2cm}
\caption{Experiments showing the impact of transferring TAGCOS-selected datasets from Llama-2 7B to Mistral-7B. Consistent improvement on TydiQA, MMLU, and BBH benchmarks demonstrate the transferability.}
\label{tab2:generalization}
\end{table}

\textbf{5\% data can achieve comparable results with full dataset.} Table~\ref{tab3:selection proportion} reveals that training with only 5\% of the data selected by TAGCOS results in performance comparable to that of the entire dataset. This can be attributed to the presence of noisy samples in the full dataset, which are less effective for fine-tuning.

\begin{table}[ht]
\centering

\begin{tabular}{l|c|c|c|c}
\hline
\textbf{prop} & \textbf{TydiQA} & \textbf{MMLU} & \textbf{BBH} & \textbf{Average} \\ \hline

\textbf{5\% }  
& 52.78  & 48.01 & 44.26  & 48.35 \\ \hline
\textbf{25\% }
& 52.13   & 49.95 & 43.33  & 48.47  \\ \hline
\textbf{100\% } 
& 51.44 & 52.96  & 44.35 & 49.58  \\ \hline
\end{tabular}
\vspace{0.2cm}
\caption{Results of experiments with different selection proportions using the Llama-2 7B model.}
\label{tab3:selection proportion}
\end{table} 

\textbf{How to determine the cluster numbers.} Table~\ref{tab4: cluster number} shows that the ideal cluster number for our setup is 100. Fewer clusters, especially less than the original dataset size of 18, fail to achieve good results. Additionally, merely increasing the number of clusters does not ensure improved performance. TAGCOS tends to degrade to plain OMP as the number of clusters increases. When the cluster count matches the number of samples, the performance is identical to plain OMP.

\begin{table}[ht]
\centering

\begin{tabular}{l|c|c|c|c}
\hline
\textbf{\# Cluster} & \textbf{TydiQA} & \textbf{MMLU} & \textbf{BBH} & \textbf{Average} \\ \hline
10  
& 54.04 & 47.71 & 40.00  & 47.25     \\ 
20
& 52.58 & 45.76 & 41.11  & 46.48  \\ 
50
& 54.84 & 47.09 & 42.96  & 48.30  \\  
100  
& 52.78  & 48.01 & 44.26  & 48.35 \\
200
& 52.57  & 46.87 & 42.87  & 47.44  \\ \hline
\end{tabular}

 \vspace{0.2cm}
\caption{Experimental results show the results on selecting different numbers of clusters. }
\label{tab4: cluster number}
\end{table} 

\textbf{Selecting early stopped checkpoints for computing gradients.} In Table~\ref{tab5: gradient checkpoint}, ``Sampled from steps before convergence'' means all the warmup checkpoint used for computing gradient features comes from the steps before convergence. ``Sampled from all training steps'' represents that these checkpoints are sampled across the entire training process evenly. We argue that ``early-selecting'', i.e., sample checkpoints from steps before convergence, works better since the gradients before convergence provide more effective reactions for data samples for training. The results in this table also support this idea. In total, it is better to have a warmup checkpoint sampled from steps before convergence to get better results on TAGCOS.

\begin{table}[ht]
\centering

\begin{tabular}{l|c|c|c|c}
\hline
  & \textbf{TydiQA} & \textbf{MMLU} & \textbf{BBH} & \textbf{Average} \\ \hline
\multicolumn{5}{c}{\textbf{Sampled from steps before convergence}} \\ \hline
  & 52.78  & 48.01 & 44.26  & 48.35 \\ \hline
\multicolumn{5}{c}{\textbf{Sampled from all training steps.}} \\ \hline
  & 53.14  & 47.16 & 39.54  & 46.61 \\ \hline
\end{tabular}
\vspace{0.2cm}
\caption{Experimental results studying the warmup checkpoint selection.}
\label{tab5: gradient checkpoint}
\end{table} 

\section{Conclusion}

This paper focuses on the effective selection of coresets for LLMs in instruction tuning. To address the challenge of accurate data representation, we utilize gradient features, which indicate the influence of each data sample on the training process. Additionally, to handle diverse collections of instruction data and ensure selection efficiency, we propose clustering similar data and applying an efficient greedy algorithm for selection. Our experimental results demonstrate the effectiveness of the entire pipeline.

\section{Limitation}

Despite its impressive performance, TAGCOS is bottlenecked by the efficiency of gradient feature estimation. The gradient feature computation stage limits its scalability to larger datasets. To effectively run TAGCOS on extensive datasets, improvements in the efficiency of gradient computation are needed.

\providecommand{\CNFX}[1]{{\em{\textrm{(#1)}}}}

\bibliographystyle{unsrtnat}
\bibliography{template}

\appendix

\section{Training Dataset Details}

In this section, we provide the detailed sources, statistics and licenses of each training dataset used in our experiment, which is shown in table~\ref{tab:dataset_details}. We conduct coreset selection from a mixture of 17 instruction tuning datasets with various scales and properties, which demonstrates superior effectiveness compared with baseline approaches.
\begin{table*}[htbp]
\centering
\begin{tabular}{l c c c}
\hline
\textbf{Dataset} & \textbf{Sourced from} & \textbf{\# Instances} & \textbf{License}\\
\hline
SuperNI & NLP datasets + Human-written Instructions & 96,913 & Apache-2.0\\
CoT & NLP datasets + Human-written CoTs & 100,000 & ODC-BY\\
Flan V2  & NLP datasets + Human-written Instructions & 100,000 & Apache-2.0\\
Dolly & Human-written from scratch & 15,011 & Apache-2.0 \\
Self-instruct & Generated w/ vanilla GPT3 LM & 82,439  & Apache-2.0\\
Unnatural Instructions & Generated w/ Davinci-002 & 68,478 & MIT\\
Code-Alpaca & Generated w/ Davinci-003 & 20,022 & Apache-2.0\\
GPT4-Alpaca & Generated w/ Davinci-003+GPT4& 52,002  & Apache-2.0\\
Baize & Generated w/ ChatGPT & 210,311 & GPL-3.0\\
ShareGPT & User prompts + outputs from various models & 168,864  & Apache-2.0 \\
WizardLM & Generated w/ GPT-3.5-Turbo	& 30,000 & - \\
Oasst1&	Human-written from scratch & 33,919 & Apache-2.0\\
Hardcoded & - & 14 & ODC-BY \\
LIMA & Human-written from scratch & 1,030 & CC-BY-NC-SA \\
Science Literature&  NLP datasets  & 7,544 & ODC-BY \\
Open-Orca & Generated w/ GPT4 & 30,000 & MIT\\
Standford Alpaca & Generated w/ Davinci-003 & 52,002 & Apache-2.0\\

\hline
\end{tabular}
\vspace{0.2cm}
\caption{Details of datasets used in our paper.}\label{tab:dataset_details}
\end{table*}

\end{document}